\def\year{2020}\relax
\documentclass[letterpaper]{article}
\usepackage{aaai20} 
\usepackage{times} 
\usepackage{helvet}
\usepackage{courier} 
\usepackage{xcolor}
\usepackage[hyphens]{url} 
\usepackage{graphicx}
\usepackage{graphicx} 
\usepackage{threeparttable}
\usepackage{booktabs}
\usepackage{amsmath}
\usepackage{amssymb}
\usepackage{dsfont}
\usepackage[linesnumbered,ruled,vlined]{algorithm2e}

\title{Reinforced Curriculum Learning on\\ Pre-trained Neural Machine Translation Models}
\author{Mingjun Zhao\textsuperscript{\rm 1}, Haijiang Wu\textsuperscript{\rm 2}, Di Niu\textsuperscript{\rm 1}, Xiaoli Wang\textsuperscript{\rm 2}\\ 
\textsuperscript{\rm 1}University of Alberta, Edmonton, AB, Canada\\
\textsuperscript{\rm 2}Platform and Content Group, Tencent, Shenzhen, China\\
zhao2@ualberta.ca, harywu@tencent.com, dniu@ualberta.ca, evexlwang@tencent.com
}
\begin{document}

\maketitle

\begin{abstract}
The competitive performance of neural machine translation (NMT) critically relies on large amounts of training data. However, acquiring high-quality translation pairs requires expert knowledge and is costly. Therefore, how to best utilize a given dataset of samples with diverse quality and characteristics becomes an important yet understudied question in NMT. Curriculum learning methods have been introduced to NMT to optimize a model's performance by prescribing the data input order, based on heuristics such as the assessment of noise and difficulty levels. However, existing methods require training from scratch, while in practice most NMT models are pre-trained on big data already. Moreover, as heuristics, they do not generalize well. In this paper, we aim to learn a curriculum for  improving a pre-trained NMT model by re-selecting influential data samples from the original training set and formulate this task as a reinforcement learning problem. Specifically, we propose a data selection framework based on Deterministic Actor-Critic, in which a critic network predicts the expected change of model performance due to a certain sample, while an actor network learns to select the best sample out of a random batch of samples presented to it. 
Experiments on several translation datasets show that our method can further improve the performance of NMT when original batch training reaches its ceiling, without using additional new training data, and significantly outperforms several strong baseline methods.
\end{abstract}

\section{INTRODUCTION}
\label{sec:intro}

Curriculum learning, as pioneered by \cite{bengio2009curriculum}, aims to improve the training of machine learning models by choosing what examples to present and in which order to present them to the learning algorithm. Curriculum learning was originally inspired by the learning experience of humans \cite{skinner1958reinforcement} \cite{peterson2004day} \cite{krueger2009flexible}---humans tend to learn better and faster when they are first introduced to simpler concepts and exploit previously learned concepts and skills to ease the learning of new abstractions. This phenomenon is widely observed in, e.g., music and sports training, academic training and pet shaping.
Without surprise, curriculum learning is found most helpful in end-to-end neural network architectures \cite{bengio2009curriculum}, since the performance that an artificial neural network can achieve critically depends on the quality of training data presented to it.   

\begin{table}[tb]
  \label{tab:sample}
  \centering
  \begin{tabular}{ll}
    \toprule
    Example 1 & \\
    zh & xianzai ta zheng kaolv zhe huijia.\\
    zh-gloss & Now he is thinking about going home.\\
    en & He is thinking about going home now.\\
    \midrule
    Example 2 & \\
    zh & wo yao chi niupai.\\
    zh-gloss & I want eat steak\\
    en & I want a steak. Get me a coke.\\
    \bottomrule
  \end{tabular}
  \caption{Examples with accurate/inaccurate translations.}
\end{table}

Neural Machine Translation \cite{kalchbrenner2013recurrent} \cite{sutskever2014sequence} (NMT) translates text from a source language to a target language in an end-to-end fashion with a single neural network. It has not only achieved state-of-the-art machine translation results, but also eliminated hand-crafted features and rules that are otherwise required by statistical machine translation. The performance of NMT has been improved significantly in recent years, as the NMT architectures evolved from the initial RNN-based models \cite{sutskever2014sequence} to convolutional seq2seq models \cite{gehring2017convolutional} and further to Transformer models \cite{vaswani2017attention}.  

However, since obtaining accurately labeled training samples in machine translation is often time-consuming and requires expert knowledge, an important question in NMT is how to best utilize a limited number of available training samples, perhaps with different lengths, qualities, and noise levels.
Recently, the application of curriculum learning is also studied for NMT. \cite{platanios2019competence} propose to feed data to an NMT model in an easy-to-difficult order and characterize the ``difficulty'' of a training example by the sentence length and the rarity of words that appear in it. Other than using the straightforward difficulty or complexity as a criterion for curriculum design, \cite{wang2018denoising} propose a method to calculate the noise level of a training example with the help of an additional trusted clean dataset and train an NMT model in a noise-annealing curriculum. 

A limitation of the existing curriculum learning methods for NMT is that they only address the batch selection issue in a ``learn-from-scratch'' scenario. Unfortunately, training an NMT model is a time-consuming task and sometimes could take up to several weeks \cite{van2017dynamic}, depending on the amount of data available. In most practical and commercial cases, a pre-trained model often already exists, while re-training it from scratch with a new ordering of batches is a waste of time and resources. In this paper, we study curriculum learning for NMT from a new perspective, that is given a pre-trained model and the dataset used to train it, to re-select a subset of useful samples from the existing dataset to further improve the model. Unlike the easy-to-difficult insights in traditional curriculum learning \cite{bengio2009curriculum}, \cite{platanios2019competence}, our idea is analogous to classroom training where a student first attends classes to learn general subjects with equal weights and then carefully reviews a subset of selected subjects to strengthen his/her weak aspects or to elevate ability in a field of interest.
 
\begin{figure}[tb]
\centering
\includegraphics[width=3.3in]{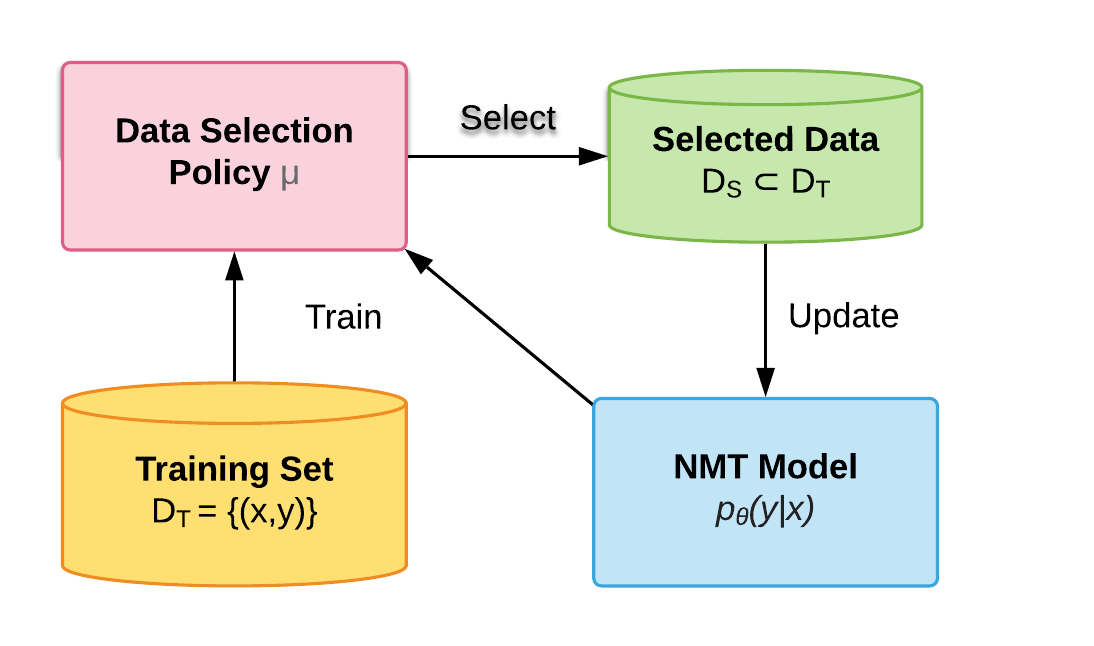}
\caption{Illustration of the curriculum learning process. The RL policy $\mu$ is used to select samples $D_S$ from the training set $D_T$. The selected data $D_S$ is used to update a pre-trained NMT model $p_\theta(y|x)$.
}
\label{fig:TaskOverview}
\end{figure}

Furthermore, while all the samples participate in batch training for the same number of epochs, it is unlikely that all data contribute equally to a best-performing model. Table~\ref{tab:sample} shows an example of two data samples from the dataset used in this paper, where Example 1 is accurately translated and can potentially improve the model better, while Example 2 is poorly translated (with unexpected words in target sentence) and may even cause performance degradation when fed to the model. The objective of our curriculum design is to identify examples from the existing dataset that may further contribute to model improvement and present them again to the NMT learning system.
An overview of our proposed task is given in Figure~\ref{fig:TaskOverview}, where useful data can be selected and fed to the system repeatedly to strengthen the model iteratively.

We formulate the data re-selection task as a reinforcement learning problem where the state is the features of $b$ randomly sampled training examples, the action is choosing one of them, and the reward is the perplexity difference on a validation set after the pre-trained model is updated with the selected sample. Thus, the primary goal of the learning problem becomes searching for a data selection policy to maximize the reward. Reinforcement learning is known to be unstable or even to diverge when the action-value function is represented by a nonlinear function, e.g., a neural network. For the sake of alleviating instability, our proposed RL framework is built based on the Deterministic Actor-Critic algorithm \cite{silver2014deterministic}. It consists of two networks, an actor network which learns a data selection policy, and a critic network which evaluates the action value of choosing each training sample while providing information to guide the training of the actor network. Besides introducing the framework, we also carefully design the state space by choosing a wide range of features to characterize each sample in multiple dimensions, including the sentence length, sentence-level log-likelihood, $n$-gram rarity, and POS and NER tagging. 

Experiments on multiple translation datasets demonstrate that our method can achieve a significant performance boost, when normal batch learning cannot improve the model further, by only re-selecting influential samples from the original dataset. Furthermore, the proposed scheme outperforms a number of other curriculum learning baseline methods, including the denoising scheme based on the use of additional trusted data \cite{wang2018denoising}.

\section{PROBLEM DEFINITION}
\label{sec:prob}

In this section, we provide a brief background of NMT and formulate the curriculum learning task on pre-trained NMT models as a reinforcement learning problem.

Machine Translation can be considered a one-to-one mapping from a source sentence $x$ to a target sentence $y$. In neural machine translation, a model parameterized by $\theta$ is searched for to maximize the conditional probability $p_\theta(y|x)$ over the training samples.
Modern NMT models adopt an encoder-decoder architecture where an encoder encodes the source sentence $x$ into a hidden representation $h$ and a decoder predicts the next target word $y_i$ given the hidden vector $h$ and all previously predicted words $\{y_1,...,y_{i-1}\}$. Thus the conditional probability is decomposed as
\begin{equation}
	\label{eq:nmtDecompose}
	\log p_\theta(y|x)=\sum_{i=1}^L \log p_\theta(y_i|y_{<i},h),
\end{equation}
where $L$ is the number of tokens in each target sentence. Given a training corpus $D_T$, the training objective of an NMT model is to minimize
\begin{equation}
	\label{eq:nmtObjective}
	J(\theta) = \sum_{(x,y)\in D_T} -\log p_\theta(y|x).
\end{equation}

We consider curriculum learning on a pre-trained NMT model, where the goal is to improve an existing model $p_\theta(y|x)$ by selecting a subset $D_S$ from the training dataset $D_T$ that led to $p_\theta(y|x)$. As compared to training from scratch, we take advantage of both the versatility of normal batch learning in the initial pre-training stage and a carefully selected curriculum for targeted model improvements. 
Specifically, our objective is to find an optimal policy $\mu_\phi$ to select $D_S$ from $D_T$ and update $p_\theta(y|x)$ with $D_S$ such that the performance of the updated model is maximized, i.e., 
\begin{equation}
\begin{aligned}
		\max_\phi &\quad \textit{perf}\Big(p_{\theta'}(y|x,\phi)\Big),\\
		\text{s.t.} & \quad p_{\theta'}(y|x,\phi) = \textit{train}\Big(p_\theta(y|x), D_S(\phi)\Big),\\
		& \quad D_S(\phi) = \mu_\phi(D_T),
\end{aligned} 
\end{equation}
where $D_S(\phi)$ is a subset selected from $D_T$ using policy $\mu_\phi$, $p_{\theta'}(y|x,\phi)$ is an updated model after training $p_\theta(y|x)$ with $D_S(\phi)$, and \textit{perf} indicates the performance of a model, e.g., measured by BLEU or Perplexity.

The main challenge is to identify and select the most beneficial data samples from $D_T$.
A naive way is to evaluate the BLEU improvement on a validation set brought by every single data sample in the training set and select the ones that improve the BLEU the most. However, this method is extremely costly and is not scalable to large datasets.

To obtain a generalizable data selection policy, we formulate the task as a reinforcement learning problem in which the environment is composed of both the dataset $D_T$ and the model $p_\theta(y|x)$. The RL agent aims to learn a policy $\mu_\phi$ which decides which sample to select when presented with a random batch of samples. 
In our framework, a state $s$ corresponds to the representation of both a data batch to select from and the NMT model, $a$ refers to the action of selecting the best data sample from the batch, and $r$ is the performance improvement of the NMT model on validation set after being updated with the selected sample. 

The RL agent is trained through interacting with the environment by repeatedly performing the following:
1) receiving a state $s$ containing a random batch of samples, 2) providing an action $a$ back to the environment according to its trained policy, and 3) updating the policy using a feedback reward $r$ given by the environment. Once the policy is trained, it can be used to select data from an arbitrarily large dataset and is scalable.

\section{METHODS}
\label{sec:model}

In this section, we describe our Deterministic Actor-Critic framework for curriculum learning, as well as the design of the state, action, and reward in detail.

\subsection{Model Overview}

For the model design of the RL agent, we choose the Deterministic Actor-Critic algorithm.

Actor-critic \cite{konda2000actor} is a widely used method in reinforcement learning combining both an actor network $\mu_\phi$ that outputs an action $a=\mu_\phi(s)$ given a state $s$ to maximize the reward, and a critic network $Q_w$ that predicts the action value $Q_w(s,a)$ of a state-action pair $(s,a)$. The actor learns a near-optimal policy via policy gradient, while the critic estimates the action-value guiding the update direction of the actor. Compared with actor-only methods like REINFORCE \cite{williams1992simple}, the existence of the critic reduces the update variance and accelerates convergence.

As our reward calculation involves evaluating the updated NMT model on a validation set and is thus expensive, we exploit a memory replay buffer to increase sample efficiency. Furthermore, we choose a deterministic policy setting as opposed to stochastic policy due to the fact that the deterministic policy gradient can be calculated much more efficiently as shown in \cite{silver2014deterministic}.

\begin{figure}[tb]
	\centering
	\includegraphics[width=3.3in]{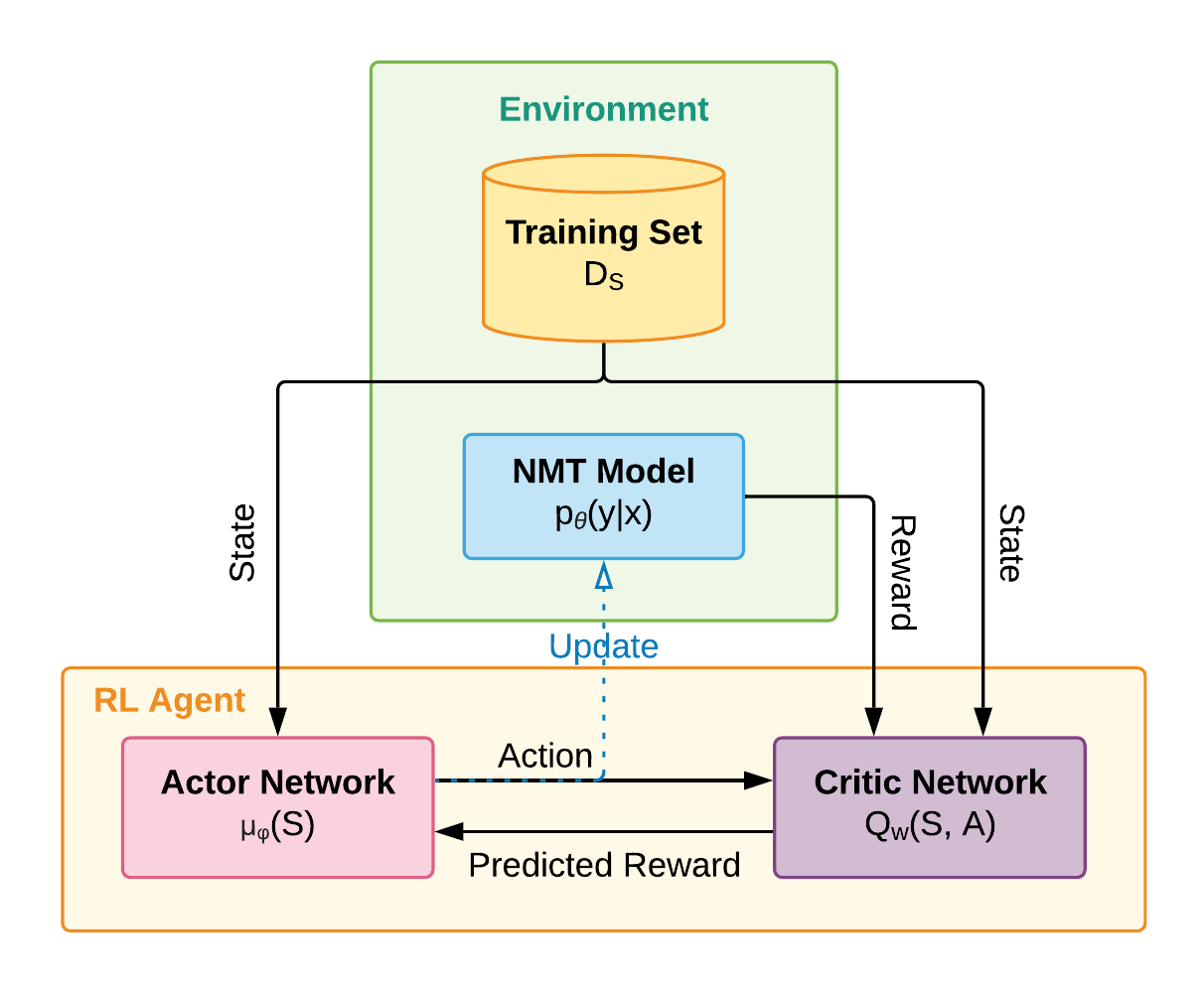}
	\caption{The proposed RL framework.}
	\label{fig:Framework}
\end{figure}

The update of the framework is illustrated in Figure~\ref{fig:Framework}. The critic network $Q_w$ takes a state-action pair $(s,a)$, evaluates the action value, and outputs a predicted reward $\tilde r =Q_w(s,a)$, and updates the parameters supervised by the actual reward $r$ from the environment. Note that the critic network provides an immediate reward per iteration. As a result, we do not need to employ additional techniques, e.g., Temporal-Difference (TD) \cite{tesauro1992practical}, to approximate the long-term reward. 
The objective of the critic network is thus to minimize the squared error of $\delta$ between the predicted reward and the actual reward, given by
\begin{equation}
	\delta_t = r_t - \tilde r_t = r_t - Q_w(s_t,a_t).
\end{equation}
The update of parameters $w$ is achieved through gradient descent as follows:
\begin{equation}
\label{eq:criticUpdate}
	w_{t+1} = w_t + \alpha_w \delta_t \nabla_w Q_w(s_t,a_t).
\end{equation}
where $\alpha_w$ corresponds to the learning rate of parameters $w$.

The actor network $\mu_\phi$ takes in a state $s$ from the environment, applies the learned policy, and outputs a corresponding action $a$. 
The objective of the actor network is to learn an optimal policy generating the proper action to maximize the predicted reward $Q_w(s,a)$. Policy gradient is used to update the parameters $\phi$, i.e.,
\begin{equation}
\label{eq:actorUpdate}
\begin{aligned}
	\phi_{t+1}&=\phi_t + \alpha_\phi \nabla_\phi Q_w(s_t,a_t)|_{a=\mu_\phi(s)}\\
	&=\phi_t + \alpha_\phi \nabla_\phi \mu_\phi(s_t) \nabla_a Q_w(s_t,a_t)|_{a=\mu_\phi(s)}.
\end{aligned}
\end{equation}

\SetKwInput{KwInput}{Input}
\SetKwInput{KwOutput}{Output}

\begin{algorithm}[tb]
\label{alg:overall}
\DontPrintSemicolon
\caption{The Proposed Method}
\KwInput{Training set $D_T$ and an NMT model $p_{\theta_0}(y|x)$}
\KwOutput{A better performing NMT model $p_{\theta_K}(y|x)$}
\For{$t=0,\ldots,K-1$}
{
	\For{\textit{number of RL training iterations}}
	{
		Sample $b$ examples from $D_T$ and form state $s$\\
		Generate action $a=\mu_\phi(s)$\\
		Compute predicted reward $\tilde r=Q_w(s,a)$\\
		Update $p_{\theta_k}(y|x)$ with selected sample to get $p_{\theta_k'}(y|x)$\\
		Calculate the perplexity difference on validation set between $p_{\theta_k}(y|x)$ and $p_{\theta_k'}(y|x)$ as reward $r$\\
		Update $Q_w$ using Eq.~(\ref{eq:criticUpdate})\\
		Update $\mu_\phi$ using in Eq.~(\ref{eq:actorUpdate})\\
	}
	Select data $D_S$ from $D_T$ using $\mu_\phi$\\
	Update $p_{\theta_k}(y|x)$ with $D_S$ to get $p_{\theta_{k+1}}(y|x)$
}

\end{algorithm}

Algorithm~\ref{alg:overall} summarizes the overall learning process of the proposed framework. In each round of data selection (with $K$ rounds in total), we first train the RL agent for an adequate number of iterations. In each iteration, we derive a state, an action, and a reward in lines 4--7 and update the actor network and the critic network in line 8 and line 9, respectively. After the RL agent is fully trained, we apply the learned policy to select a subset $D_S$ and use $D_S$ to update the NMT model $p_{\theta_k}(y|x)$ and move to the next round. Usually one or two rounds are sufficient.

\begin{figure}[tb]
\centering
\includegraphics[width=3.3in]{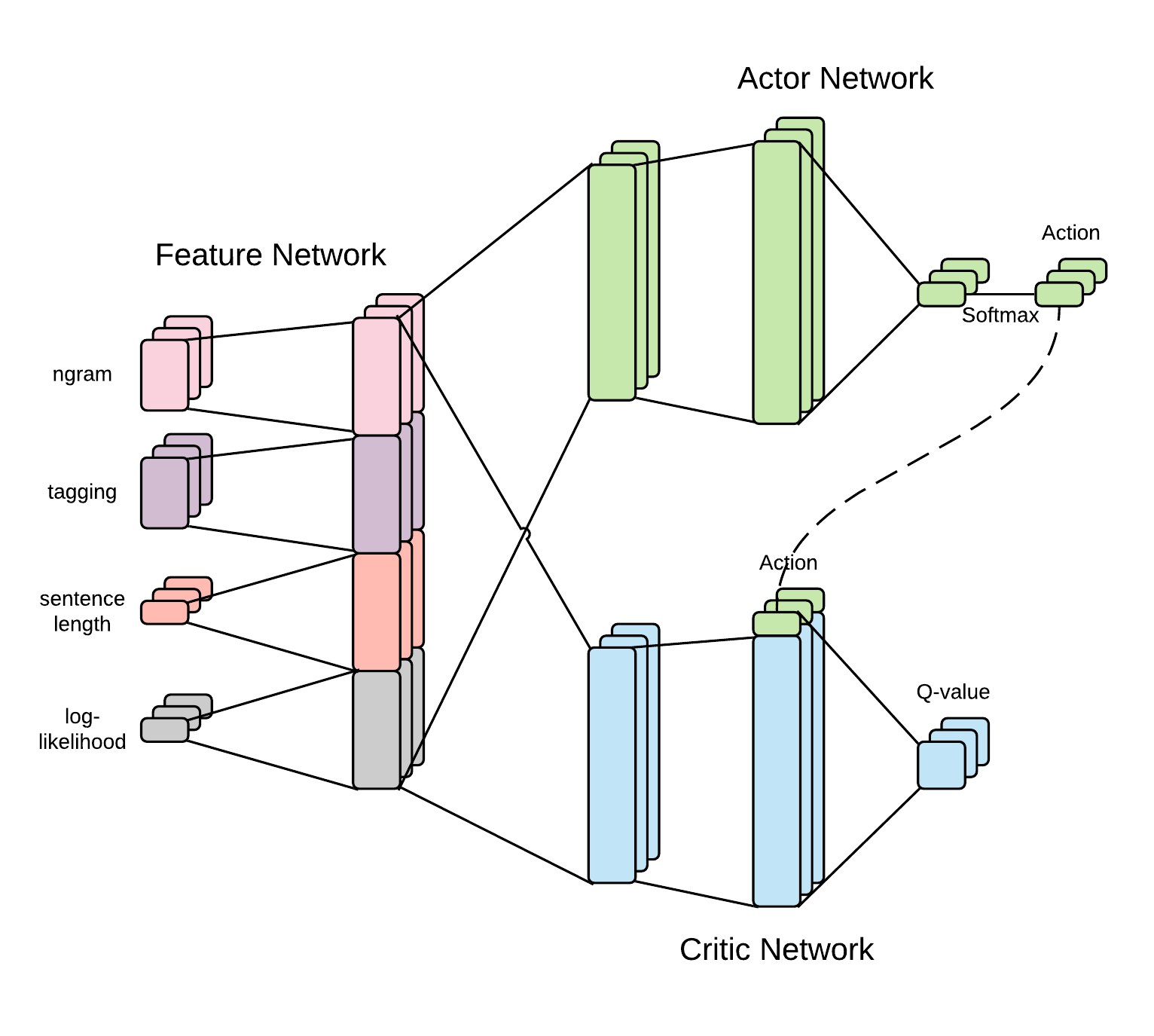}
\caption{The network structure of the RL agent, where the depth represents a batch (of 3 samples in this illustration).
}
\label{fig:Network} 
\end{figure}

Figure~\ref{fig:Network} demonstrates the network structure of the RL agent. A \emph{feature network} is shared between the Actor Network $\mu_\phi$ and the Critic Network $Q_w$. It takes in the raw features of the sampled batch of $b$ examples, where each feature of each sample goes through an independent single-layer MLP. The concatenation of the outputs constitutes the state representation $s$. Note that different examples in the sampled batch share the same network weights.

The Actor Network $\mu_\phi$ is composed of a two-layer MLP and computes a score for each example in the sampled batch of $b$ examples based on the input state $s$, and outputs the action $a$ as a probability vector representing the probability of each example being selected, by taking a softmax operation over the computed scores of $b$ examples.

The Critic Network $Q_w$ also has two layers and calculates the action value of a given state-action pair $(s,a)$, where the action $a$ is the output of the Actor Network, i.e., $a=\mu_\phi(s)$, and is concatenated to the second layer of the critic network.

Note that although the feature network is shared between both the actor network $\mu_\phi$ and the critic network $Q_w$, we only update it with the critic network to reduce training instability.

\subsection{State}
\label{subsec:state}

The state $s$ is meant to be a full summarization of the environment including a data batch of $b$ examples to select from and information about the pre-trained model. However, the number of parameters in the pre-trained model is too large to be included in the state directly at each time step. Thus, we need to find a representation that can represent both the samples to be selected and the existing model using a limited number of parameters. In our state design, we focus on three different dimensions, namely \emph{informativeness}, \emph{uncertainty}, and \emph{diversity}. We use the sentence length as a measure for informativeness, the sentence-level log-likelihood for uncertainty, and the $n$-gram rarity together with NER and POS taggings for diversity.

The most intuitive representation feature of a parallel sentence is the sentence length, i.e. the number of tokens in a sentence. This simple scalar roughly measures the amount of information involved in a sentence and is also used in \cite{platanios2019competence} as a ``difficulty'' estimate. 

Following the intuition that examples yielding large uncertainty can benefit the model performance \cite{li2006confidence}, another feature we use is the sentence-level log-likelihood $L_{p_\theta(y|x)}$ calculated by the pre-trained model $p_\theta(y|x)$ by summing up the log probability of each word $y_i$ of the target sentence:
\begin{equation}
	\label{eq:logp}
	L_{p_{\theta}(y|x)} = \log p_\theta(y|x) = \sum_{y_i \in Y} \log p_\theta(y_i|x).
\end{equation}

\cite{yang2015multi} and \cite{sener2018active} suggest that selecting samples that are farthest away from the existing dataset can benefit model training. 
Incorporating this idea, we further utilize two other feature vectors, $n$-gram rarity and taggings, to represent the similarity between a given sample and the entire training set. 

For $n$-gram rarity, we use $n=\{1,2,3,4\}$. Given all the sentences in $D_T$, we define the relative frequency for a unique $n$-gram $g^{(n)}_j$ in $D_T$ as
\begin{equation}
	\label{eq:ngram}
	f^{(n)}(g^{(n)}_j) \triangleq \frac{1}{N^{(n)}} \sum_{s_k \in D_T} \sum_{g^{(n)}_i \in g^{(n)}_{(s_k)}} \mathds{1}_{g^{(n)}_i=g^{(n)}_j},
\end{equation}
where $j=1,...,\#\textit{\{unique n-grams\}}$, $s_k$ is a sentence from $D_T$, $g^{(n)}_{(s_k)}$ is all $n$-grams in $s_k$, and $N^{(n)}$ is the total number of $n$-grams in $D_T$. For $n$-gram representation of a sentence $s_k$, we form all the $n$-gram frequencies into a vector:
\begin{equation}
	\label{eq:ngramSent}
	F^{(n)}(s_k) = \{f(g^{(n)}_1),...,f(g^{(n)}_L)|g^{(n)}_l \in g^{(n)}_{(s_k)}\}.
\end{equation}
In this paper, we use $n=\{1,2,3,4\}$ and calculate the $n$-gram vectors for both the source and target sentences.

For taggings, we use Named Entity Recognition (NER) and Parts of Speech (POS) and apply ideas similar to our $n$-gram design. As the tagging of a word is dependent of the sentence that the word lies in, a same word may be tagged differently in various sentences. To give an example in POS tagging, the word ``Book'' can either be tagged as ``NOUN'' or ``VERB'' depending on its meaning in the sentence. If most occurrences of the word ``Book'' are tagged as ``NOUN'' in the training set, the model may not be able to correctly learn the second meaning of it. Therefore, the model should be fed more with samples in which ``Book'' is tagged as ``VERB''. In order to reflect this phenomenon, we define a tagging value for a word $w$ and its current tag $t$ as
\begin{equation}
	\label{eq:tagging}
	v(w,t) \triangleq \frac{1}{N_w} N_{tag(w)=t},
\end{equation}
where $N_w$ is the number of times the word $w$ has appeared in the training set and $N_{tag(w)=t}$ is the number of times the word $w$ is tagged with tag $t$ among all its occurrences. 
Similarly, for both NER and POS taggings, the tagging values of words form a vector representation of a sentence:
\begin{equation}
	\label{eq:taggingSent}
	V(s_k) = \Big\{v\Big(w_1,t_1\Big),...,v\Big(w_L,t_L\Big)|w_l \in s_k\Big\}.
\end{equation}

\subsection{Action and Reward}
\label{subsec:actionReward}

In the task of curriculum learning, an action represents the process of data selection. \cite{fang2017learning} assume a stream-based setting where data examples come one by one in a stream, and design their action as making a decision on whether or not a single incoming data example should be selected. We argue that our problem setting is actually \emph{pool-based} instead of stream-based, where a pool of data exists for selection and deciding on the selection of each individual sample per step would be inefficient. In \cite{kumar2019reinforcement}, a dataset is split into several bins according to a noise measure of data samples, and the action determines from which bin the next batch of training data should be selected. However, this method highly depends on an effective heuristic criterion for bin formation and is thus hard to generalize. 

Therefore, we propose our action design which samples a batch of $b$ data examples from $D_T$, computes a score for each example according to the trained policy, and choose the one with the highest score. Correspondingly, in our design, we can easily control the size of $D_S$ by varying the batch size $b$, i.e., $|D_S| = |D_T|/b$, since we choose one out of $b$ samples for each batch.

For the choice of reward signal, we use the performance improvement of the NMT model evaluated on the validation set after it is updated with the selected sample. \emph{Perplexity} is used as the performance metric instead of BLEU as it is shown to be more consistent and less noisy \cite{so2019evolved}. We assign a reward of $0$ to unselected samples. 

\section{EXPERIMENTS}
\label{sec:simu}

In this section, we will first describe the datasets used in our evaluation and provide the implementation details along with the performance results.

\subsection{Datasets \& Metrics}
To compare our proposed method with other curriculum learning methods on NMT task, we conduct comprehensive empirical analysis on several zh-en translation datasets:

\begin{itemize}
	\item{\textbf{MTD}} is an internal news translation dataset with 1M samples in the training set and 1,892 samples in both the validation set and the test set. The average length of source sentences is $19.55$ and the average length of target sentences is $21.06$. 
	\item{\textbf{CASICTB, CASIA2015, NEU}} are three independent datasets with 1M, 2M, and 2M examples from different data sources in WMT18 which is a public translation dataset in news area with more than 20M samples. We only use a part of data from WMT18 to evaluate our method. All three datasets share the same validation set \emph{newsdev2017} and the test set \emph{newstest2017} both composed of 2k samples.
\end{itemize}

\begin{table}[tb]
  \resizebox{\columnwidth}{!}
  {
    \begin{threeparttable}
      \caption{Description of evaluation datasets.}
      \label{tab:datasets}
      \begin{tabular}{llllll}
        \toprule
        Dataset & Train & Val & Test & Src-Len & Tgt-Len\\
        \midrule
        MTD &  $1,000,000$ & $1,892$ & $1,892$ & $19.55$ & $21.06$\\
        CASICTB &  $994,248$ & $2,002$ & $2,001$  & $22.61$ & $23.95$\\
        CASIA2015 &  $1,049,988$ & $2,002$ & $2,001$  & $23.89$ & $25.19$\\
        NEU &  $1,999,946$ & $2,002$ & $2,001$  & $16.95$ & $18.54$\\
        \bottomrule
      \end{tabular}
    \end{threeparttable}
  }
\end{table}

Table~\ref{tab:datasets} summarizes the details of the experimental datasets. The columns titled ``Train'', ``Val'', and ``Test'' correspond to the number of examples in training, validation, and test sets. ``Src-Len'' and ``Tgt-Len'' correspond to the average sentence length of source language and target language respectively.

For evaluation metrics, we use Perplexity for the reward calculation in RL agent training, and report BLEU \cite{papineni2002bleu} for the final performance of the NMT models.

\subsection{Experimental Settings}
We implement our models in PyTorch 1.1.0 \cite{paszke2017pytorch} and train the model with a single Tesla P40. We utilize NLTK \cite{bird2009natural} to perform POS and NER taggings.

For the NMT model, we use the OpenNMT implementation \cite{opennmt} of Transformer \cite{vaswani2017attention}. It consists of a 6-layer encoder and decoder, with 8 attention heads, and 2,048 units for the feed-forward layers. The multi-head attention model dimension and the word embedding size are both set to 512. During training, we use Adam optimizer \cite{kingma2015adam} with a learning rate of 2.0 decaying with a noam scheduler and a warm-up steps of 8,000. Each training batch contains 4,096 tokens and is selected with bucketing \cite{kocmi2017curriculum}. During inference, we employ beam search with a beam size of 5. 

For the RL agent, we use an Deterministic Actor-Critic architecture and build our system based on \cite{deeprl}. In our framework, we use several tricks proven to be effective to RL training including a memory replay buffer of size 2,500, a warm-up phase of 500 steps, and a target network which is updated by mixing weights with the on-line network with a mix factor of $0.1$. For calculating rewards, we train the NMT model with the single selected sample using SGD and a learning rate of 1e-4.

The feature network maps data features of sentence length, sentence-level log-likelihood, taggings, and $n$-gram rarity to vectors of size 1, 8, 16, and 32 respectively with a FC layer, and concatenates them together as a shared state representation. The actor network is composed of two FC layers with hidden sizes of 300 and 400. The critic network is designed the same as the actor network except that the output action from actor network is concatenated to the second layer. Relu is used as the activation function in each FC layer in this network.

In experiments, we conduct two rounds of RL agent training and data selection. In each round, the RL agent is trained for 20k steps and the best model with the highest sum of rewards during the last 1,000 steps is used for data selection. The RL agent keeps selecting data given randomly sampled batches from the training set $D_T$ and feed the selected data to the NMT model until no performance improvement is observed. For the training process on selected data, we keep the NMT model's optimizer and learning rate setting unchanged.

We use different batch sizes of the sampled batch $b$ for the two rounds of training and selection with $b_1=16$ and $b_2=128$ indicating in the first round we select 1 sample from 16 and in the second round, from 128 samples we select the best one. This is because we think in order to achieve improvement further on the basis of the first round, a stricter data selection criterion must be applied.

\subsection{Baselines}
To make comparisons with other existing curriculum methods, we have conducted several baseline experiments. 

We take the core ideas of existing curriculum learning methods of training on data samples with gradually increasing difficulty \cite{platanios2019competence} and gradually decreasing noise \cite{wang2018denoising} and apply them to our setting with pre-trained models. We evaluate the following three baseline methods along with our proposed method.

\begin{itemize}
	\item{\textbf{Denoising}} is a curriculum learning method of training an NMT model in a noise-annealing fashion \cite{wang2018denoising}. They propose to measure NMT data noise with the help of a trusted dataset which contains generally cleaner data compared to the training dataset. \cite{kumar2019reinforcement} also utilize data noise in their curriculum design and achieve similar performance as \cite{wang2018denoising}. For the choice of the trusted dataset, we choose a subset of 500 sentences from the validation set newsdev2017 of CASICTB, CASIA2015 and NEU.
	\item{\textbf{Sentence Length}} is an intuitive difficulty measure used in \cite{platanios2019competence}, since longer sentences tend to contain more information and more complicated sentence structure. 
	\item{\textbf{Word Rarity}} is another metric for measuring the sample difficulty, as rare words appear less frequently in the training process and should be presented to the learning system more. The formula for calculating the word rarity of a sentence can be found in \cite{platanios2019competence}.
\end{itemize}

For baseline experiments, the pre-trained NMT model is further trained on 20\% of the original data, which are selected by the above criteria, i.e., the least noisy, the longest, and the highest word rarity, respectively.

\subsection{Main Results}

\begin{table*}[tb]
	\centering
    \begin{threeparttable}
      \caption{Performance comparison of our proposed method with other baseline methods using BLEU on different datasets.}
      \label{tab:results}
      \begin{tabular}{l|cccc}
        \toprule
        Method & MTD & CASICTB & CASIA2015 & NEU\\
        \midrule
        Base &  $18.21$ $(+0.00)$ & $13.43$ $(+0.00)$ & $18.65$ $(+0.00)$ & $20.06$ $(+0.00)$\\
        Sentence Length &  $18.34$ $(+0.13)$ & $13.52$ $(+0.09)$ & $18.75$ $(+0.10)$ & $20.21$ $(+0.15)$\\
        Word Rarity &  $18.35$ $(+0.14)$ & $13.49$ $(+0.06)$ & $18.72$ $(+0.07)$ & $20.17$ $(+0.11)$\\
        Denoising &  $18.42$ $(+0.21)$ & $13.55$ $(+0.12)$ & $18.86$ $(+0.21)$ & $20.44$ $(+0.38)$\\
        Ours-1 Round &  $18.81$ $(+0.60)$ & $13.60$ $(+0.17)$ & $18.91$ $(+0.26)$ & $20.38$ $(+0.32)$\\
        \midrule
        Ours-2 Rounds &  $\textbf{19.11}$ $(+\textbf{0.90})$ & $\textbf{14.03}$ $(+\textbf{0.60})$ & $\textbf{19.24}$ $(+\textbf{0.59})$ & $\textbf{20.79}$ $(+\textbf{0.73})$\\
        \bottomrule
      \end{tabular}
    \end{threeparttable}
\end{table*}

Table~\ref{tab:results} compares the performance of our method with other baseline methods on different datasets evaluated using BLEU. The result shows that our proposed method significantly out-performs other baseline methods by a great margin. We conduct two rounds of training and update in our experiments. While the result of the first round surpasses almost all the baseline methods, our second round further improves the performance and achieves a final BLEU improvement of $+0.90$, $+0.60$, $+0.59$, and $+0.71$ on MTD, CASICTB, CASIA2015, and NEU respectively over the pre-trained model.

The reason of our success is due to our utilization of an RL framework to proactively select data samples that are potentially beneficial to the training of the NMT model. First, we formulate the task of curriculum learning on pre-trained NMT models as a reinforcement learning problem. Second, we construct an effective design of state, action and reward. Our state representation includes features of different dimensions of informativeness, uncertainty and diversity. Third, we propose a Deterministic Actor-Critic framework that learns a policy to select the best samples from the training set to improve the pre-trained model. By incorporating all these designs together, our proposed framework is able to achieve a significant performance enhancement on the pre-trained NMT model.

\subsection{Analysis}
We evaluate the impact of different modules and methods by ablation test on MTD dataset. Table~\ref{tab:ablation} list the performance of our model variants with different features included.

We incrementally accommodate different features of examples to the state by first starting from sentence length and sentence-level log-likelihood as they are both scalars. The performance increased slightly by $0.26$ BLEU compared with the pre-trained base model. Then we further accommodate $n$-gram rarity and POS and NER taggings to the state vectors, and observe a larger increase of performance of $0.42$ and $0.60$ respectively. Finally, we incorporate the second round of RL agent training and data selection on the basis of the result of first round, and achieve the best performance with a $0.90$ BLEU increase. Note that a stricter selection policy is applied to the second round (128 choose 1) compared with the first round (16 choose 1). 
	
\begin{table}[tb]
	\centering
    \begin{threeparttable}
      \caption{Ablation analysis on different features of our proposed framework on MTD.}
      \label{tab:ablation}
      \begin{tabular}{l|l}
        \toprule
        Method & MTD\\
        \midrule
        Base & $18.21$ $(+0.00)$\\
        Senlen + Logp &  $18.47$ $(+0.26)$\\
        + N-gram &  $18.63$ $(+0.42)$\\
        + Tagging &  $18.81$ $(+0.60)$\\
        + 2nd Round &  $\textbf{19.11}$  $(+\textbf{0.90})$\\
        \bottomrule
      \end{tabular}
    \end{threeparttable}
\end{table}

\section{RELATED WORK}
\label{sec:related}

High-quality machine translation corpus is costly and difficult to collect, thus it is necessary to make the best use of the corpus at hand. A straightforward method to achieve this goal is to remove the noisy samples in the training data, and train an new model with the clean ones. Unfortunately, it is hard to estimate the quality of a parallel sentence in the absence of golden reference \cite{fan2019bilingual}.  Moreover, \cite{wang2018denoising} find that some of the noisy samples may yield some benefits to the model performance. Then they define a new method of computing noise level of a data example with the help of an extra trusted dataset and propose to train an NMT model in a noise-annealing curriculum.

Curriculum learning aims to organize the training process in a meaningful way by feeding certain samples to the model in certain training stage such that the model can learn better and faster \cite{bengio2009curriculum}. They propose a simple strategy which organizes all training samples into bins of similar complexity and starts training from the easiest bin to include more complexed bins until all bins are covered. \cite{kocmi2017curriculum} apply this idea to NMT by binning according to simple features like sentence length and word frequency, and improve this strategy by restricting that each sample can only be trained once during an epoch. \cite{zhang2018empirical} conduct empirical studies on several hand-crafted curriculum and adopt a probabilistic view of curriculum learning. \cite{platanios2019competence} further propose a competence function $c(t)$ with respect to training time step $t$ as the indicator of learning progress and select samples based on both difficulty and competence. However, these heuristic-based approaches highly depend on hand-crafted curriculum and are hard to generalize.

Compared with heuristic-based approaches, RL-based policy learning models are trained end-to-end and do not rely on hand-crafted strategies. \cite{tsvetkov2016learning} use Bayesian optimization to learn a linear model for ranking examples in a work-embedding task. \cite{graves2017automated} explore bandit optimization for scheduling tasks in a multi-task problem.\cite{wu2018reinforced} select examples in a co-trained classifier using RL. \cite{kumar2019reinforcement} organize the dataset into bins based on the data noise proposed by \cite{wang2018denoising} and utilize a DQN to learn a data selection policy deciding from which bin to select the next batch. 

Different from the existing curriculum learning methods, our work focuses on learning a training curriculum with reinforcement learning for an existing pre-trained NMT model. We argue that existing curriculum learning methods are only applicable on train-from-scratch scenarios, and learning from an existing model can save training time.

Active learning \cite{settles2009active} is another related area which focuses on selectively obtaining labels for unlabeled data in order to improve the model with least labeling cost. \cite{haffari2009active} \cite{bloodgood2010bucking} study active learning for Statistical Machine Translation using some hard-coded heuristics. \cite{fang2017learning} design an active learning algorithm based on a deep Q-network, in which the action corresponds to binary annotation decisions applied to a stream of data. \cite{liu2018learning} make use of imitation learning to train a data selection policy.

\section{CONCLUSION}
\label{sec:conclude}

In this paper, we study curriculum learning for NMT from a new perspective, to re-select a subset of useful samples from the existing dataset to further improve a pre-trained model, and formulate this task as a reinforcement learning problem. Compared with existing curriculum methods only applicable on train-from-scratch scenarios, our setting saves training time by better utilizing the existing pre-trained models. Our proposed framework is built based on the deterministic actor-critic algorithm, and learns a policy to select examples that can improve the model the most. 
We conduct experiments on several zh-en translation datasets and compare our method with other baseline methods including the easy-to-difficult curriculum and the denoising scheme. 
Through rounds of training and data selection, our method achieves a significant performance boost on the pre-trained model, and out-performs all baselines methods by a great margin. 

\fontsize{9.0pt}{10.0pt}\selectfont
\bibliographystyle{aaai}
\bibliography{main}

\end{document}